\documentclass[10pt,twocolumn,letterpaper]{article}

\usepackage[pagenumbers]{wacv} 

\usepackage{graphicx}
\usepackage{amsmath}
\usepackage{amssymb}
\usepackage{booktabs}
\usepackage{times}
\usepackage{epsfig}
\usepackage{graphicx}
\usepackage{amsmath}
\usepackage{amssymb}
\usepackage{booktabs}

\usepackage{microtype}
\usepackage{algorithm2e}
\usepackage{color}
\usepackage{colortbl}
\usepackage{multirow}
\usepackage{xcolor}
\usepackage[tableposition=top]{caption}
\usepackage{arydshln}
\usepackage{float}
\usepackage{bbding}
\RestyleAlgo{ruled}

\usepackage[pagebackref,breaklinks,colorlinks]{hyperref}
\usepackage{orcidlink}

\usepackage[capitalize]{cleveref}
\crefname{section}{Sec.}{Secs.}
\Crefname{section}{Section}{Sections}
\Crefname{table}{Table}{Tables}
\crefname{table}{Tab.}{Tabs.}

\def\wacvPaperID{***} 
\def\confName{WACV}
\def\confYear{2024}

\begin{document}

\title{\textit{What's in the Flow?} Exploiting Temporal Motion Cues for Unsupervised Generic Event Boundary Detection}

\author{
Sourabh Vasant Gothe \orcidlink{0000-0003-4737-2218},\quad
Vibhav Agarwal \orcidlink{0000-0002-2029-9885},\quad
Sourav Ghosh \orcidlink{0000-0003-1866-1408},\quad
Jayesh Rajkumar Vachhani \orcidlink{0000-0003-0267-4474},\\
Pranay Kashyap,\quad
Barath Raj Kandur Raja \orcidlink{0000-0003-0451-2452}\\[0.2cm]
\textit{Samsung R\&D Institute Bangalore, India}\\
{\tt\small \{sourab.gothe, vibhav.a, sourav.ghosh, jay.vachhani, kashyap.p, barathraj.kr\} @samsung.com}
}

\maketitle

\begin{abstract}
Generic Event Boundary Detection (GEBD) task aims to recognize generic, taxonomy-free boundaries that segment a video into meaningful events. Current methods typically involve a neural model trained on a large volume of data, demanding substantial computational power and storage space. We explore two pivotal questions pertaining to GEBD: Can non-parametric algorithms outperform unsupervised neural methods? Does motion information alone suffice for high performance? This inquiry drives us to algorithmically harness motion cues for identifying generic event boundaries in videos. In this work, we propose FlowGEBD, a non-parametric, unsupervised technique for GEBD. Our approach entails two algorithms utilizing optical flow: (i) Pixel Tracking and (ii) Flow Normalization. By conducting thorough experimentation on the challenging Kinetics-GEBD and TAPOS datasets, our results establish FlowGEBD as the new state-of-the-art (SOTA) among unsupervised methods. FlowGEBD exceeds the neural models on the Kinetics-GEBD dataset by obtaining an F1@0.05 score of 0.713 with an absolute gain of 31.7\% compared to the unsupervised baseline and achieves an average F1 score of 0.623 on the TAPOS validation dataset.

\end{abstract}

\section{Introduction}

In 2023, video has accounted for 82.5\% of all web traffic, making it the most popular form of content on the internet. Increased video consumption has made video understanding a critical task in computer vision, comprising video classification \cite{wang2018temporal, feichtenhofer2020x3d, liu2022video}, object segmentation \cite{xu2021vitae, xu2022towards}, action localization \cite{chao2018rethinking, wang2021proposal, zhang2022actionformer}, and captioning \cite{zhu2020actbert, wang2022geb+}, among others. However, the memory requirements of the models, the feasibility of real-time inference, and domain generalization are typical constraints on these solutions.\looseness=-1

\begin{figure}[t]
	\centering
	\includegraphics[width=0.85\linewidth]{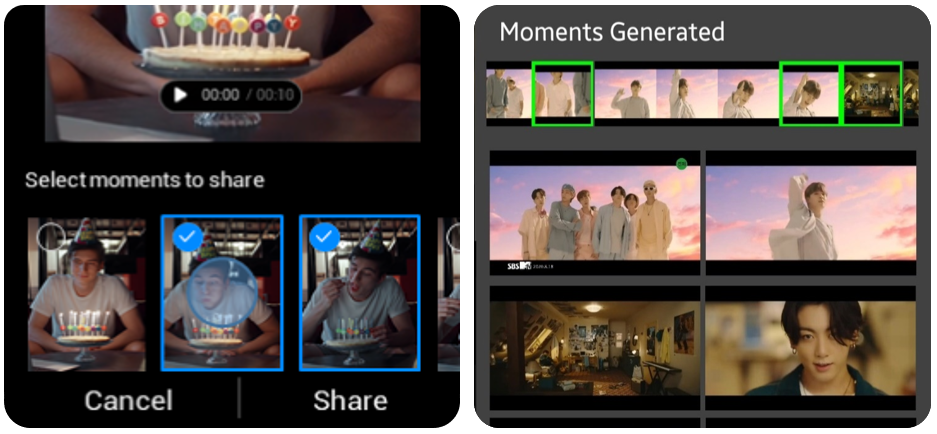}
    \caption{FlowGEBD enables applications on smartphones, like short video segment sharing, summarization, editing by identifying generic video moments}
    \vspace{-2pt}
	\label{fig:usecase}
\end{figure}

Current state-of-the-art video models~\cite{simonyan2014two, tran2018closer, feichtenhofer2019slowfast, kondratyuk2021movinets, li2022mvitv2} have been mainly focused on building upon a limited set of predefined action classes and usually process short clips to generate global video-level predictions. With the growth of video content, the number of classes is expanding, and the predefined target classes cannot encompass them all. Recently, the GEBD \cite{shou2021generic} task was introduced with the objective of studying the long-form video understanding problem through the lens of human perception~\cite{tversky2013event}. GEBD aims to locate class-agnostic event boundaries in a video, regardless of its category. It considers the following high-level causes of event boundaries: changes in subject, action, shot, environment, and object of interaction. The outcome of GEBD has many potential applications: video summarization, video editing, short video segment sharing (Fig.~\ref{fig:usecase}), and enhancing video classification and other downstream tasks~\cite{wang2022geb+}.\looseness=-1

\setlength{\textfloatsep}{0.3cm}
\begin{figure}[t]
	\centering
	\includegraphics[width=\linewidth,height=0.65\linewidth]{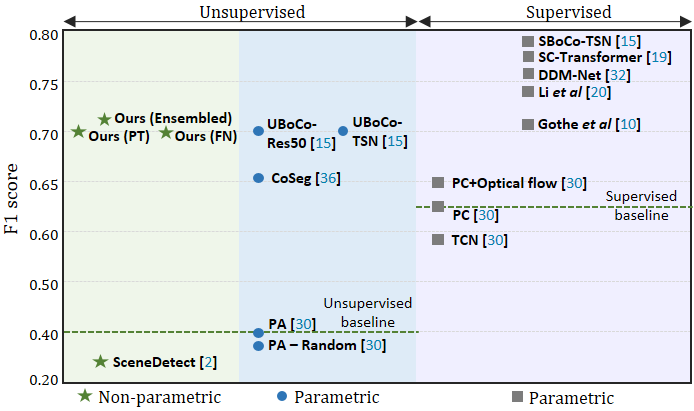}
    \caption{F1@0.05 scores of different methods on the Kinetics-GEBD validation dataset. Our method FlowGEBD achieves state-of-the-art results among unsupervised methods compared to non-parametric~\cite{castellano2018pyscenedetect} and parametric~\cite{ shou2021generic, kang2022uboco, wang2021coseg} benchmarks. \looseness=-1}
	\label{fig:teaser}
\end{figure}

There are three main approaches to the GEBD problem: supervised, unsupervised, and self-supervised. Several supervised methods \cite{kang2021winning, tang2022progressive, hong2021generic, li2022structured, li2022end, gothe2023self} propose to decode the self-similarity representation produced by the frame features for boundary detection. The literature investigates a wide variety of approaches, including efficient representation learning \cite{kang2021winning, gothe2023self}, transformer decoders \cite{li2022structured}, attention masks \cite{tang2022progressive}, and the use of compressed domain features \cite{li2022end}. In unsupervised approaches,  CoSeg \cite{wang2021coseg} investigates cognitively inspired parametric methods, and UBoCo \cite{kang2022uboco} yields the best results by employing a novel contrastive loss. In self-supervised approaches, TeG \cite{qian2021exploring} investigates learning temporal granularity in video representations. In contrast, Rai \etal \cite{rai2023motion} employ a differentiable motion feature learning module to detect spatial and temporal differences for GEBD. These methods incorporate explicit motion features into their network structures and earn a high F1 score as they are deep neural networks (DNN) guided by the ground truth.
However, we approach GEBD from a different perspective, investigating a two-fold question:
(a) Can non-parametric algorithms outperform unsupervised parametric methods?
(b) Is motion information alone sufficient to achieve high performance? We seek answers to these questions by exploiting the motion information to detect the generic event boundaries in a video algorithmically.\looseness=-1

In this paper, we present two unsupervised, non-parametric approaches to solve GEBD. (i) Pixel Tracking (PT) method that relies on sparse optical flow in temporal dimension to identify the boundary, and (ii) Flow normalization (FN) method that traces the max temporal dense flow to detect the event boundaries. The ensemble of both achieves an F1 score of 0.713 on Kinetics-GEBD and an F1 score of 0.375 on the TAPOS. 

As shown in Fig.  \ref{fig:teaser}, our method achieves 31.7\% absolute gain compared to the unsupervised baseline method and outperforms the supervised baseline \cite{shou2021generic} by 8.8\% on the Kinetics-GEBD dataset. In summary, our main contributions are as follows:\looseness=-1
\begin{itemize}
\item We propose FlowGEBD, a non-parametric (algorithmic), unsupervised method for generic event boundary detection.\looseness=-1

\item We design two algorithms by leveraging motion information: (i) Pixel Tracking (PT) and (ii) Flow-Normalization (FN) using optical flow estimation in framewise and patchwise mode to solve the GEBD task.\looseness=-1 

\item We conduct extensive ablations, time complexity analysis, and sensitivity analysis to demonstrate the robustness of the proposed method.\looseness=-1

\item Our results establish FlowGEBD as the new state-of-the-art among unsupervised methods on the challenging Kinetics-GEBD and TAPOS datasets.\looseness=-1

\end{itemize}

\vspace{-0.4em}
\section{Related Work}

\subsection{Generic Event Boundary Detection}
\vspace{-0.3em}
Generic event boundary detection (GEBD)~\cite{shou2021generic} aims to localize the moments where humans naturally perceive taxonomy-free event boundaries that break a longer video into shorter temporal segments. Previous methods~\cite{shou2021generic, tang2022progressive, hong2021generic} formulate the GEBD task as binary classification, which predicts the boundary label of each frame by considering the temporal context information. However, it could be more efficient because the redundant computation is conducted while generating the representations of consecutive frames. Kang \etal~\cite{kang2021winning} proposed to use the temporal self-similarity matrix (TSM) as intermediate representation and used contrastive learning as an auxiliary to learn better from the TSM results. Li \etal~\cite{li2022end} proposed solving GEBD using the compressed video features and achieved \begin{math}4.5\times\end{math} faster-running speed than the baseline method~\cite{shou2021generic} on GPU. Recently, Gothe \etal \cite{gothe2023self} developed the most miniature model to solve the GEBD task with the lowest inference time on GPU. SC-Transformer~\cite{li2022structured} introduced a structured partition of sequences (SPoS) mechanism to learn structured context using a transformer-based architecture for GEBD. To enrich motion information, optical flow is introduced as a new modality in~\cite{https://doi.org/10.48550/arxiv.2206.12634}. TeG~\cite{qian2021exploring} proposed a generic self-supervised model for learning persistent and more fine-grained features and uses a 3D-ResNet-50 encoder as its backbone. However, all these methods require substantial memory and computational resources along with the labeled data.\looseness=-1

Regarding unsupervised GEBD approaches, PySceneDetect~\cite{castellano2018pyscenedetect} is a Python library that detects shot changes by considering pixel changes in the HSV colorspace. However, generic event boundaries consist of various boundary causes like the change of action, subject, and environment, implying that only a tiny portion of event boundaries can be detected with this approach. PredictAbility (PA)~\cite{shou2021generic} computationally assesses the predictability score over time and then locates the event boundaries by detecting the local minima of the predictability sequence. CoSeg \cite{wang2021coseg} devises a transformer-based frame feature reconstruction scheme and adopts ResNet-18~\cite{he2016deep} as the backbone. UBoCo~\cite{kang2022uboco} proposes an unsupervised/supervised method using the TSM as the video representation. UBoCo's unsupervised framework for GEBD combines Recursive TSM Parsing (RTP) and the Boundary Contrastive (BoCo) loss. However these models belong to a high memory regime. \looseness=-1

\subsection{Learning motion and visual correspondences}
Motion plays a crucial role in video understanding, and many SOTA models ~\cite{liu2019learning, kwon2020motionsqueeze, https://doi.org/10.48550/arxiv.2206.12634, rai2023motion, tang2022progressive} incorporate motion information by using optical flows. Lucas and Kanade's image registration method~\cite{lucas1981iterative}, also known as gradient-based optical flow, enables motion estimation possible with high-speed computation. Pyramidal Lucas and Kanade~\cite{bouguet2001pyramidal}, Gunnar Farneback~\cite{farneback2003two, farneback2002polynomial, farneback2000fast} are other well-known methods for motion estimation.\looseness=-1

DDM-Net~\cite{tang2022progressive} applies progressive attention to multilevel dense difference maps (DDM) to characterize motion patterns and jointly aggregate motion and appearance cues in a supervised setting. MotionSqueeze (MS)~\cite{kwon2020motionsqueeze} introduces an end-to-end trainable, model-agnostic and lightweight module to extract motion features on the fly for video understanding. However, it requires training via backpropagation and integration with pre-existing video architectures. Rai \etal~\cite{rai2023motion} presents a self-supervised model for GEBD by reformulating training objectives at frame-level and clip-level to learn effective video representations using the MS~\cite{kwon2020motionsqueeze} module. However, these are parametric methods that require training on large datasets. To the best of our knowledge, there is no unsupervised and non-parametric (algorithmic) solution with high performance in generic event boundary detection.\looseness=-1

\vspace{-0.5em}
\section{Proposed Methodology}
GEBD takes a video as input and returns a set of boundary timestamps. Mathematically, it maps an ordered sequence of $L$ frames, $\left<f_1, f_2, \dots, f_L\right>$ (that may also be represented as $\overrightarrow{F} \in \mathbf{F}$), to a set of timestamps $\left\{b_1, b_2, \dots, b_M\right\}$ ($=\mathcal{B} \in \mathbf{B}$), that denote the event boundaries. It then naturally follows that $M \le (L-1)$. For all practical purposes, $M \ll L$ and $\forall b_i \in \mathcal{B}, \exists j$, such that timestamp $b_i$ corresponds to a unique frame $f_j$. Thus, we formulate the GEBD task as:
\begin{equation}
    \mathcal{T}: \mathbf{F} \rightarrow \mathbf{B}
    \vspace{-0.8em}
\end{equation}

\begin{figure}[tbp]
	\centering
	\includegraphics[width=\linewidth]{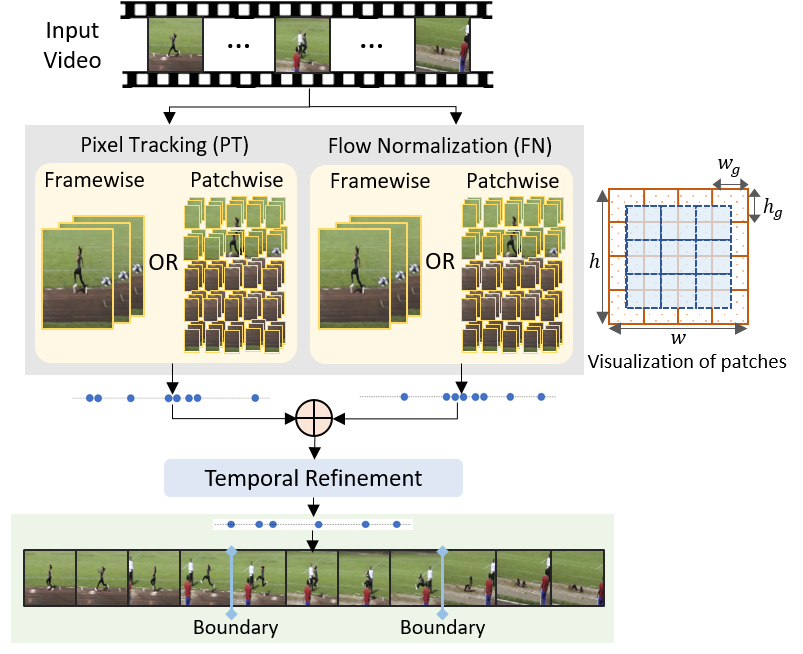}
    \caption{FlowGEBD accepts a video as input and predicts a set of event boundaries, $\mathcal{B}$. Visual representation of patches with $n_w=n_h=4$ (right). {\color[rgb]{.8,.4,0} $\Box$}: Base patches, {\color[rgb]{0,.1,.9} $\Box$}: Centroidal}
	\label{fig:flowchart}
\end{figure}

Here, we describe our approach, FlowGEBD, that solves this task using pixel tracking, flow normalization, and their ensemble (with temporal refinement) as shown in Fig. \ref{fig:flowchart}.\looseness=-1

\subsection{FlowGEBD with Pixel Tracking (PT)} \label{sec:PT}

In this section, we present a method that leverages sparse optical flow to determine event boundaries by monitoring the flow of a subset of pixels.

\vspace{-1em}
\subsubsection{Framewise mode}\label{sec:PT_framewiseFlowGEBD}
\vspace{-0.4em}
We process a video frame-by-frame, considering each frame as a unit. Each frame $f$ of width $w$ and height $h$ comprises a 2-dimensional matrix of pixels, $p_{u,v}$, where $u,v \in \mathbb{Z^{+}}$ (positive integers), $u \in [1, w]$, and $v \in [1, h]$. We only consider the luminance information of pixels. Hence, $p_{u,v}$ can be represented as a real number ($p_{u,v} \in \mathbb{R}$), $0 \le p_{u,v} \le 1$.\looseness=-1

The apparent motion of pixel $p_{u,v}$ between two consecutive frames caused by the movement of an object or camera is measured by optical flow.
For each frame $f_i$ with a subsequent frame $f_{i+1}$, the optical flow $\Phi_i$ can be denoted as a 2-dimensional matrix of displacement vectors \cite{simonyan2014two}. Each element in the displacement vector $\overrightarrow{d}_{u,v}$ denotes the horizontal and vertical motion of $p_{u,v}$ between frames $f_i$ and $f_{i+1}$.\looseness=-1

\begin{algorithm}[t]
	\caption{FlowGEBD using Pixel Tracking (Framewise mode)}
	\label{algo:PixelTracking}
	\KwData{Video of resolution $w \times h$ as a sequence of $L$ frames, $\overrightarrow{F} = \left<f_1, f_2, \dots, f_L\right>$}
    \KwResult{Event Boundaries, $\mathcal{B} = \left\{b_1, b_2, \dots, b_M\right\}$}
    \hrule
    \hfill\\
    
    $\mathcal{B}_{\text{temp}} \gets \{\}$

	$\mathcal{P}_{1} \gets \mathcal{P}_{\text{base}} \gets \text{samplePixels}(f_1)$ 
	
	{\color{gray} \scriptsize \hfill// $O(1)$ for uniform random}

	\For{$f_i \in \left(\overrightarrow{F} - f_1\right)$}{
		
		$\Phi_{i-1} \gets \text{sparseFlow}(f_{i-1}, f_i, \mathcal{P}_{i-1})$
		
		{\color{gray} \scriptsize \hfill// $O\left(wh\left|\mathcal{P}_{\text{base}}\right|\right)$}
		\begin{flalign*}
		    \mathcal{P}_{i} &\gets \bigcup_{\overrightarrow{d}_{u,v} \in \Phi_{i-1}}^{\left|\overrightarrow{d}_{u,v}\right| \neq 0} \Bigl\{p_{u,v}\Bigr\} && \text{\color{gray} \scriptsize // Non-zero flow, $O(wh)$}
		\end{flalign*}
		\If{ $\frac{\left|\mathcal{P}_{i}\right|}{\left|\mathcal{P}_{\text{base}}\right|} < \theta_1$}{
		    \hfill\\
			
			$\mathcal{B}_{\text{temp}} \gets \mathcal{B}_{\text{temp}} \cup \{i\}$
			
			$\mathcal{P}_{i} \gets \mathcal{P}_{\text{base}} \gets \text{samplePixels}(f_i)$ 
		}
	}{\color{gray} \scriptsize \hfill// \textbf{for}-loop $\implies O\left(Lwh\left|\mathcal{P}_{\text{base}}\right|\right)$} \\
	$\mathcal{B} \gets \text{refine}(\mathcal{B}_{\text{temp}})$    {\color{gray} \scriptsize \hfill// Refer to Algorithm \ref{algo:filterBoundaries}}

	\Return $\mathcal{B}$
	\[\]\hrule\hfill\\

    {
        \scriptsize
        $\text{samplePixels}(\cdot)$: Uniform random or Shi-Tomasi corner detection
        
        $\text{sparseFlow}(\cdot)$: Sparse optical flow using iterative Lucas-Kanade method
        
        $\theta_1$: A constant threshold 
    }
\end{algorithm}

\vspace{-1em}
\paragraph{Method.} The key intuition is that an event boundary can be determined by monitoring the optical flow of a subset of pixels. This underlying assumption is supported by Shou et al.~\cite{shou2021generic} who consider change in brightness, rapid camera movements, etc. as definitive indicators of event boundaries.
So, for the first frame $f_1$ in the sequence $\overrightarrow{F}$, we use uniform random sampling or Shi-Tomasi corner detection algorithm~\cite{shi1994tomasi} to identify a set $\mathcal{P}_{\text{base}}$ comprised of key features (pixels $p_{u,v}$). Then, for every subsequent frame $f_i$, we compute the sparse flow for these pixels using the iterative Lucas-Kanade method~\cite{lucas1981iterative}. We consider a pixel as also an element of the current key pixel set $\mathcal{P}_{\text{current}}$ (= $\mathcal{P}_{\text{i}}$), if and only if it has non-zero displacement $\overrightarrow{d}_{u,v}$ from the previous frame.\looseness=-1

In each of these frame-by-frame iterations, whenever the ratio of elements in $\mathcal{P}_{\text{current}}$ to $\mathcal{P}_{\text{base}}$ falls below a predefined threshold $\theta_1$, we infer to have encountered an event boundary and record the current frame index. In such a scenario, we resample new key pixels $\mathcal{P}_{\text{base}}$ from the current frame. If no such event boundary is encountered in an iteration, we maintain $\mathcal{P}_{\text{base}}$ as a constant reference until a boundary is identified. To determine the final set of boundaries, we apply a temporal boundary refinement algorithm (Section \ref{sec:ensemble}). At any time, this algorithm depends only on the past frame to deduce an event boundary, thereby exhibiting the causal property. The overall approach is detailed in Algorithm \ref{algo:PixelTracking}.\looseness=-1

\vspace{-1em}
\paragraph{Can we improve further?} The framewise approach monitors the key pixels in a frame. However, for a video of a moderately large field of view, it is often the case that an event boundary may be denoted by the change in actions of certain subjects in the video, even if the background remains static. Furthermore, the main subjects of the frame are typically positioned along the grid lines and at the intersections to make it more aesthetically pleasing following the Rule of thirds~\cite{dhar2011high}. For such cases, the event boundary can be determined more accurately if we decompose a frame into a grid of patches instead of having the entire frame as a single unit. Patchwise processing provides advantages like capturing subject change in small areas and detecting action change from one patch to another. Thus, in the next section, we propose an approach that processes each frame as a composition of multiple patches.\looseness=-1
\vspace{-1em}
\subsubsection{Patchwise mode}\label{sec:PT_patchwiseFlowGEBD}

\vspace{-0.5em}
A patch $g_f$, derived from frame $f$, consists of a subset of the frame pixels. More specifically, patch $g_f(u,v,w_g,h_g)$ consists of all pixels, $p_{i,j} \in f$, where $i,j \in \mathbb{Z^{+}}$, $i \in \left[u, u + w_g\right)$ and $j \in \left[v, v + h_g\right)$. We denote the set of all such patches in frame $f$ as $\mathbf{G}_{f}$.

We define two categories of patches where patches of the same category do not overlap each other. The first among these are \textit{``base patches''}, which distribute the pixels equally and independently along the width and height of the frame. We refer to the other category as \textit{``centroidal patches''}, as each of their edges joins the centroids of adjacent base patches. Centroidal patches help capture events that span across the intersections of base patches. Fig.  \ref{fig:flowchart} (right)  depicts the arrangement of patches, where the total number of patches (each of width $w_g$ and height $h_g$) is given by:        
\begin{equation}
    \mathcal{N}_g = \underbrace{n_w\times n_h}_{\text{Base patches}} + \underbrace{(n_w-1)\times(n_h-1)}_{\text{Centroidal patches}}
\end{equation}
Where $n_w$ and $n_h$ represent the cardinality of base patches along the frame width and height, respectively. 

\vspace{-0.4em}
\paragraph{Method.} In this mode, we independently process the entire frame sequence $\mathcal{N}_g$ times, once for each patch, considering only the corresponding patches. During this, we skip the temporal boundary refinement stage mentioned in Algorithm \ref{algo:PixelTracking}. We then take a union of the $\mathcal{N}_g$ predicted boundary sets and apply boundary refinement to derive $\mathcal{B}$. It may be noted that framewise mode is a specialized case of patchwise, where $n_w = n_h = 1 \implies \mathcal{N}_g = 1$.

\subsection{FlowGEBD with Optical Flow Normalization}
We explore another approach that leverages dense optical flow and determines event boundaries by observing the normalized optical flow.
\vspace{-1em}
\subsubsection{Framewise mode}
\vspace{-0.5em}
As we have observed in FlowGEBD with pixel tracking (section \ref{sec:PT_patchwiseFlowGEBD}), the framewise mode is a specialized case of patchwise. Hence, we discuss the Optical Flow Normalization method in the more generalized patchwise mode, and the same can be adapted for framewise by using $n_w = n_h = 1$.
\vspace{-1em}
\subsubsection{Patchwise mode}

\paragraph{Method.} For every frame $f_i$, we identify the set of $\mathcal{N}_g$ patches, denoted by $\mathbf{G}_{f_i}$, for a fixed $n_w$ and $n_h$. Then, for every consecutive frame $f_{i-1}$ and $f_i$, we compute $\mathcal{N}_g$ dense optical flows (one between each patch pair, $g_{f_{i-1}}$ and $g_{f_i}$). For this, we use the Gunnar Farneback algorithm \cite{farneback2003two}. Then, we use the maximum flow displacement corresponding to all patch pixels as the \textit{``flow of the patch''} or its \textit{``PatchFlow''}. After processing $L$ frames and their $\mathcal{N}_g$ patches, we accumulate the PatchFlow for each patch across temporal dimension and normalize the displacement values. We hypothesize that considerable displacement of PatchFlow in the temporal dimension constitutes a change in action or event. For any patch, if the normalized value for frame index $i$ exceeds a constant threshold $\theta_2$, we deem to have encountered an event boundary and add the corresponding frame index $i$ to our working set of event boundaries $\mathcal{B}_{\text{temp}}$. Finally, we apply temporal refinement on $\mathcal{B}_{\text{temp}}$ to compute the refined event boundary set $\mathcal{B}$. This approach for patchwise processing of dense optical flow is detailed in Algorithm \ref{algo:FlowBased}.

\begin{algorithm}[tb!]
	\caption{FlowGEBD using Optical Flow Normalization (Patchwise mode)}
	\label{algo:FlowBased}
	\KwData{Video of resolution $w \times h$ as a sequence of $L$ frames, $\overrightarrow{F} = \left<f_1, f_2, \dots, f_L\right>$; Parameters: Patch width $w_g$, and height $h_g$}
    \KwResult{Event Boundaries, $\mathcal{B} = \left\{b_1, b_2, \dots, b_M\right\}$}
	\hrule
	\hfill\\
	
	$\mathbf{G}_{f_1} \gets \text{patches}(f_1, w_g, h_g)$    {\color{gray} \scriptsize \hfill// Gets $\mathcal{N}_g$ patches (Fig. \ref{fig:flowchart})}
	
	\For{$f_i \in \left(\overrightarrow{F} - f_1\right)$}{
		$\mathbf{G}_{f_i} \gets \text{patches}(f_i, w_g, h_g)$
		
		$\mathbf{\Phi}^g_{i-1} \gets \{\}$  {\color{gray} \scriptsize \hfill// Placeholder for all optical flows in frame $f_i$}

		\For{\footnotesize $\left(g_{f_{i-1}}, g_{f_i}\right) \in \left(\mathbf{G}_{f_{i-1}}, \mathbf{G}_{f_i} | \text{\scriptsize $\genfrac{}{}{0pt}{}{\text{corresponding}}{\text{patches}}$}\right)$}{
	        \hfill\\
		    
			$\varphi \gets \text{denseFlow}(g_{f_{i-1}}, g_{f_i})$ {\color{gray} \scriptsize \hfill// $O(w_g h_g)$)}

			$\mathbf{\Phi}^g_{i-1} \gets \mathbf{\Phi}^g_{i-1} \cup \left\{\max\left(\varphi\right)\right\}$
		}	{\color{gray} \scriptsize \hfill// inner-\textbf{for}-loop $\implies O(\mathcal{N}_g w_g h_g)$)}
		
	}   {\color{gray} \scriptsize \hfill// outer-\textbf{for}-loop $\implies O(L\mathcal{N}_g w_g h_g)$)}
	
	{\color{gray} \scriptsize \hfill// We now have \textit{PatchFlows} $= \{\mathbf{\Phi}^g_{1}, \mathbf{\Phi}^g_{2}, \dots, \mathbf{\Phi}^g_{L-1}\}, \dots$}
	
	{\color{gray} \scriptsize \hfill// $\dots$where $i^{\text{th}}$ element = set $\mathbf{\Phi}^g_{i}$ of patch optical flows for $f_i$}

	$\mathcal{B}_{\text{temp}} \gets \{\}$

	\For{\footnotesize $\left(\Phi_1^g, \Phi_2^g, \dots, \Phi_{L-1}^g\right) \in \text{PatchFlows } |$ {\scriptsize $\genfrac{}{}{0pt}{}{\text{corresponding}}{\text{patches}}$}}{
	    $\overrightarrow{\Phi} \gets \left(\Phi_1^g, \Phi_2^g, \dots, \Phi_{L-1}^g\right)$
	    
	   $\hat{\Phi} \gets \frac{\overrightarrow{\Phi}}{||\overrightarrow{\Phi}||_2}$ {\color{gray} \scriptsize \hfill// L2-norm}

	    $\mathcal{B}_{\text{temp}} \gets \mathcal{B}_{\text{temp}} \cup \left\{
	        \arg_{i}^{\hat{\Phi}_i > \theta_2}\left(\hat{\Phi}\right) \right\}$
	}   {\color{gray} \scriptsize \hfill// \textbf{for}-loop $\implies O(L\mathcal{N}_g)$)}
	
	$\mathcal{B} \gets \text{refine}(\mathcal{B}_{\text{temp}})$    {\color{gray} \scriptsize \hfill// Refer to Algorithm \ref{algo:filterBoundaries}}
	
	\Return $\mathcal{B}$
	\[\]\hrule\hfill\\
	{
	    \scriptsize   
	    $\text{denseFlow}(\cdot)$: Dense optical flow using the Gunnar Farneback's algorithm
	    
	    $\theta_2$: A constant threshold
	}
	
\end{algorithm}

\subsection{FlowGEBD with ensembling of Pixel Tracking and Flow Normalization} \label{sec:ensemble}

Pixel tracking helps determine event boundaries based on the sparse optical flow of a few key pixels. On the other hand, flow normalization aggregates the dense optical flow of all pixels and offers a lossless method to determine a set of event boundaries using ``PatchFlow''. To obtain an ensemble of both approaches, we independently take the event boundaries from both without performing the temporal refinement stage. Instead, we take a union of the predicted sets from these two approaches and perform temporal refinement over the union.

\begin{algorithm}[btp]
	\caption{Temporal Refinement of Boundaries}
	\label{algo:filterBoundaries}
	\KwData{Event Boundaries, $\mathcal{B} = \left\{b_1, b_2, \dots, b_M\right\}$}
    \KwResult{Refined Event Boundaries, $\mathcal{\tilde{B}} \subseteq \mathcal{B}$}
    \hrule
    \hfill\\
    
    $\mathcal{\tilde{B}}_{\text{temp}} \gets \{\}; $ $\overrightarrow{\mathcal{B}} \gets \text{sorted}(\mathcal{B})$ {\color{gray} \scriptsize \hfill// $O(M \log M)$}
    
    $\overrightarrow{K} \gets \left<\right>$ {\color{gray} \scriptsize \hfill// Placeholder for a cluster of boundary elements}
    
    \For{$b^{'}_i\in \overrightarrow{\mathcal{B}}$}{
                
        \If{ $\left(\overrightarrow{K} \ne \left<\right>\right) \land \left(\forall k \in \overrightarrow{K} \mid \left(\left|b^{'}_i - k\right| \ge \theta_3\right)\right)$}{

            $\mathcal{\tilde{B}}_{\text{temp}} \gets \mathcal{\tilde{B}}_{\text{temp}} \cup \left\{\text{median}\left(\overrightarrow{K}\right)\right\}$
            
            $\overrightarrow{K} \gets \left<\right>$
        }
        
        $\overrightarrow{K} \gets \overrightarrow{K}^\frown \left<b^{'}_i\right>$
    }

    $\mathcal{\tilde{B}}_{\text{temp}} \gets \mathcal{\tilde{B}}_{\text{temp}} \cup \left\{\text{median}\left(\overrightarrow{K}\right)\right\}$ {\color{gray} \scriptsize \hfill// Flush $\overrightarrow{K}$}

    \Return $\mathcal{\tilde{B}}_{\text{temp}}$ \textbf{as} $\mathcal{\tilde{B}}$
	\hfill\\\hrule\hfill\\
	{
	    \scriptsize
	    $\theta_3$: A constant threshold
	}
    
\end{algorithm}

\setlength{\textfloatsep}{0.05cm}
\begin{table*}[t]
	\begin{center}
		\resizebox{0.8\textwidth}{!}{%
			\begin{tabular}{c|c|cccccccccc|c}
				\hline
				\textbf{Supervision}                            & {\textbf{Rel.Dis. threshold}} & \textbf{0.05}  & \textbf{0.1}   & \textbf{0.15}  & \textbf{0.2}   & \textbf{0.25}  & \textbf{0.30}  & \textbf{0.35}  & \textbf{0.4}   & \textbf{0.45}  & \textbf{0.5}   & \textbf{Avg}   \\ \hline
				\multirow{8}{*}{Supervised }                            & PC~\cite{shou2021generic} (Baseline) & 0.625  & 0.758   & 0.804  & 0.829   & 0.844  & 0.853  & 0.859  & 0.864   & 0.867  & 0.870   & 0.817   \\ 
				& PC + Optical Flow~\cite{li2022end} & 0.646 & 0.776 & 0.818 & 0.842 & 0.856 & 0.864 & 0.868 & 0.874 & 0.877 & 0.879 & 0.830 \\
				& Gothe \etal \cite{gothe2023self} & 0.712 & - & - & - & - & - & - & - & - & - & - \\
                & SBoCo-Res50 \cite{kang2022uboco} & 0.732 & - & - & - & - & - & - & - & - & - & 0.866 \\
                & DDM-Net \cite{tang2022progressive} & 0.764 & 0.843 & 0.866 & 0.880 & 0.887 & 0.892 & 0.895 & 0.898 & 0.900 & 0.902 & 0.873 \\
                & Li \etal \cite{li2022end} & 0.743 & 0.830 & 0.857 & 0.872 & 0.880 & 0.886 & 0.890 & 0.893 & 0.896 & 0.898 & 0.865 \\
                & SC-Transformer \cite{li2022structured} & 0.777 & 0.849 & 0.873 & 0.886 & 0.895 & 0.900 & 0.904 & 0.907 & 0.909 & 0.911 & 0.881 \\
                & SBoCo-TSN \cite{kang2022uboco} & \textbf{0.787} & - & - & - & - & - & - & - & - & - & 0.892 \\
				\hline
				\multirow{10}{*}{Unsupervised} & PA - Random~\cite{shou2021generic}$^\dagger$       & 0.336 & 0.435 & 0.484 & 0.512 & 0.529 & 0.541  & 0.548  & 0.554  & 0.558  & 0.561  & 0.506  \\
				& PA~\cite{shou2021generic}$^\dagger$                & 0.396 & 0.488 & 0.520 & 0.534 & 0.544 & 0.550  & 0.555  & 0.558  & 0.561  & 0.564  & 0.527  \\ 
				 & CoSeg~\cite{wang2021coseg}$^\dagger$                &  0.656 & 0.758  & 0.783 &  0.794  & 0.799  & 0.803  & 0.804  & 0.806  & 0.807  & 0.809  & 0.782  \\
				 & UBoCo-Res50~\cite{kang2022uboco}$^\dagger$          & 0.703 & 0.839 & 0.862 & 0.885 & 0.889 & 0.893 & 0.894 & 0.898 & 0.900 & 0.902 & 0.866  \\
				& UBoCo-TSN~\cite{kang2022uboco}$^\dagger$        & 0.702 & 0.846 & 0.862 & 0.879 & 0.888 & 0.889 & 0.895 & 0.897 & 0.904 & 0.905 & 0.866 \\ 
				& SceneDetect~\cite{castellano2018pyscenedetect}$^\star$                             & 0.275 & 0.300 & 0.312 & 0.319 & 0.324 & 0.327 & 0.330 & 0.332 & 0.334 & 0.335 & 0.318 \\
				& Ours (PT \ref{algo:PixelTracking})$^\star$      &  0.702 & 0.819 & 0.844 & 0.855 & 0.860 & 0.863 & 0.866 & 0.867 & 0.869 & 0.870 & 0.841    \\
				& Ours (FN \ref{algo:FlowBased})$^\star$         & 0.691 & 0.826 & 0.860 & 0.877 & 0.885 & 0.889 & 0.892 & 0.894 & 0.896 & 0.897 & {0.861}  \\
				& Ours (Ensembled)$^\star$ & \textbf{0.713} & 0.828 & 0.850 & 0.858 & 0.862 & 0.864 & 0.866 & 0.867 & 0.868 & 0.869 & 0.845\\ \hline
			\end{tabular}%
		}
	\end{center}
	\caption{F1 results on Kinetics-GEBD validation set with different Rel.Dis. thresholds. FlowGEBD achieves the best F1@0.05 scores for unsupervised setting (31.7\% absolute gain over unsupervised baseline, PA~\cite{shou2021generic}). $\dagger$ : parametric (neural) Methods $\star$ : Non-parametric}

	\label{tab:KINETICS_RESULTS}
\end{table*}

\vspace{-1.7em}
\paragraph{Temporal refinement.} We analyze the elements of a set of predicted boundary timestamps along the corresponding temporal dimension to identify ``rare boundaries'' and ``popular boundaries''. Rare or isolated boundaries are those where the event changes in a single patch and typically with no neighboring timestamps, i.e., for a rare boundary $b_r$, there exists no other boundary within the duration $b_r \pm \theta_3$. On the other hand, popular boundaries are dense clusters where multiple boundaries have been determined within the temporal vicinity.\looseness=-1

We may interpret contiguous popular boundaries as belonging to one cluster and each rare boundary as a standalone single-element cluster. Then, we select one representative element for each cluster by identifying its median boundary element and consider only such elements for the final set of refined event boundaries. The generic Algorithm \ref{algo:filterBoundaries} identifies such clusters and determines an optimal boundary in each of them.\looseness=-1

\setlength{\textfloatsep}{0.1cm}
\setlength{\floatsep}{0.1cm}
\begin{table*}[t]
	\begin{center}
		\resizebox{0.8\textwidth}{!}{%
			\begin{tabular}{c|c|cccccccccc|c}
				\hline
				\textbf{Supervision}                            & {\textbf{Rel.Dis. threshold}} & \textbf{0.05}  & \textbf{0.1}   & \textbf{0.15}  & \textbf{0.2}   & \textbf{0.25}  & \textbf{0.30}  & \textbf{0.35}  & \textbf{0.4}   & \textbf{0.45}  & \textbf{0.5}   & \textbf{Avg}   \\ \hline
				\multirow{3}{*}{Supervised}
                & PC \cite{shou2021generic} & 0.522 & 0.595 & 0.628 & 0.647 & 0.660 & 0.666 & 0.672 & 0.676 & 0.680 & 0.684 & 0.643 \\
                & DDM-Net \cite{tang2022progressive} & 0.604 & 0.681 & 0.715 & 0.735 & 0.747 & 0.753 & 0.757 & 0.760 & 0.763 & 0.767 & 0.728 \\
                & SC-Transformer \cite{li2022structured} & \textbf{0.618} & 0.694 & 0.728 & 0.749 & 0.761 & 0.767 & 0.771 & 0.774 & 0.777 & 0.780 & 0.742 \\
                \hline
				\multirow{6}{*}{Unsupervised}
				& PA - Random~\cite{shou2021generic}$^\dagger$      & 0.158 & 0.233 & 0.273 & 0.310 & 0.331 & 0.347  & 0.357  & 0.369  & 0.376  & 0.384  & 0.314  \\
				& PA~\cite{shou2021generic}$^\dagger$               & 0.360 & 0.459 & 0.507 & 0.543 & 0.567 & 0.579 & 0.592 & 0.601 & 0.609 & 0.615 & 0.543  \\ 

				& SceneDetect~\cite{castellano2018pyscenedetect}$^\star$                             & 0.035 & 0.045 & 0.047 & 0.051 & 0.053 & 0.054 & 0.055 & 0.056 & 0.057 & 0.058 & 0.051 \\
				& Ours (PT \ref{algo:PixelTracking})$^\star$       & 0.355 & 0.489 & 0.562 & 0.619 & 0.655 & 0.677 & 0.693 & 0.703 & 0.714 & 0.721 & 0.619 \\
				& Ours (FN \ref{algo:FlowBased})$^\star$       & 0.346 & 0.487 & 0.562 & 0.619 & 0.658 & {0.678} & 0.695 & {0.706} & {0.715} & {0.722} & 0.618  \\
				& Ours (Ensembled)$^\star$ & \textbf{0.375} & {0.502} & {0.569} & {0.624} & {0.658} & 0.677 & {0.695} & 0.703 & 0.711 & 0.717 & {0.623} \\ \hline
			\end{tabular}%
		}
	\end{center}
	\caption{F1 results on TAPOS validation set with different Rel.Dis. thresholds. The ensembled method achieves the best F1 score compared to other unsupervised methods. $\dagger$ : parametric (neural) Methods $\star$ : Non-parametric Methods}
	\label{tab:TAPOS_RESULTS}
	\vspace{-2em}
\end{table*}

\section{Experiments}
In this section, we conduct multiple experiments and evaluate both algorithms, followed by the ensembled method.

\subsection{Dataset}\label{sec:dataset} 

\paragraph{Kinetics-GEBD.} Our approach is evaluated primarily on the challenging Kinetics-GEBD \cite{shou2021generic}, a benchmark dataset for locating the boundaries of generic events in the video. It consists of 54,691 videos of 10 seconds each that span a broad spectrum of video domains in the wild and is open-vocabulary, taxonomy-free. The ratio of the train, validation, and test sets in Kinetics-GEBD is equal, with each set including roughly 18,000 videos chosen from Kinetics-400 \cite{kay2017kinetics}. FlowGEBD is an algorithmic unsupervised method, so we do not require train data. We evaluate our methods on the validation set, as the annotations of the test sets are not public.\looseness=-1

\vspace{-1em}
\paragraph{TAPOS.} In addition to Kinetics-GEBD, we experiment on the TAPOS dataset \cite{shao2020intra} containing Olympics sports videos with 21 actions. The dataset authors manually defined how to break each action into sub-actions during annotation. Following \cite{shou2021generic}, we re-purpose TAPOS for the GEBD task by performing boundaries localization between sub-actions in each action instance. TAPOS contains 1790 instances for validation, and we evaluate on the same.\looseness=-1

\vspace{-0.6em}
\paragraph{Implementation and Evaluation.} We run all our experiments on Intel(R) Xeon(R) Silver 4210 CPU @ 2.20GHz equipped machine. We sample the video at 4 FPS and resize it to $160\times160$ as preprocessing. As described in \cite{shou2021generic}, we use F1 at 0.05 Relative Distance (Rel.Dis.) as our primary evaluation metric. The predicted boundary is deemed accurate for a certain Rel.Dis. threshold if the difference between the predicted and ground truth timestamps is smaller than the threshold. We report F1 scores of different thresholds to range from 0.05 to 0.5 with a gap of 0.05.\looseness=-1

\vspace{-0.3em}
\subsection{Main Results}

\paragraph{Kinetics-GEBD.}
Table \ref{tab:KINETICS_RESULTS} illustrates the results of our methods on the Kinetics-GEBD validation set along with unsupervised and supervised benchmarks. PT achieves a higher F1@0.05 of 0.702 than FN, with a strong recall of 0.91 (average) and a 0.77 (average) precision. Intuitively, PT is able to detect action change across patches along with subject change as the cardinality of the patch increases. FN obtains a high Avg. F1 value through a balanced precision and recall of 0.81 and 0.90, respectively. The FlowGEBD (Ensembled) outperforms all previous unsupervised methods with the highest F1@0.05 of 0.713 using a refinement approach (Algorithm  \ref{algo:filterBoundaries}) that combines PT and FN. Compared to unsupervised baseline PA~\cite{shou2021generic}, FlowGEBD obtains a significant gain of 31.7\% in F1@0.05 and exceeds DNN-based unsupervised methods~\cite{wang2021coseg,kang2022uboco}, demonstrating the effectiveness of our proposed algorithms. Additionally, compared to the supervised baselines PC~\cite{shou2021generic} and PC + Optical Flow~\cite{li2022end}, our method achieves 8.8\% and 6.7\% absolute improvement, respectively.\looseness=-1

\vspace{-1.0em}
\paragraph{TAPOS.}
We also conduct experiments on the TAPOS~\cite{ shao2020intra}; the results are summarized in Table \ref{tab:TAPOS_RESULTS}. The dataset is not inherently well-suited for GEBD as it comprises a pre-defined set of 21 action classes. Hence, we separate sub-action instances from each action video and treat them as a single video for GEBD. Shou \etal\cite{shou2021generic} have shown that the GEBD model trained on TAPOS underperforms on the Kinetics-GEBD dataset due to a change in boundary semantics. However, our algorithm is robust enough to be applied directly to the TAPOS dataset. Compared to the unsupervised benchmark PA~\cite{shou2021generic}, our method obtains Avg. F1 score of 0.623, gaining 8\% absolute improvement. We found no alternative SOTA unsupervised methods for GEBD on the TAPOS dataset to compare our results directly.\looseness=-1

\setlength{\textfloatsep}{0.02cm}
\begin{figure}[bt]
	\centering
	\includegraphics[width=0.8\linewidth]{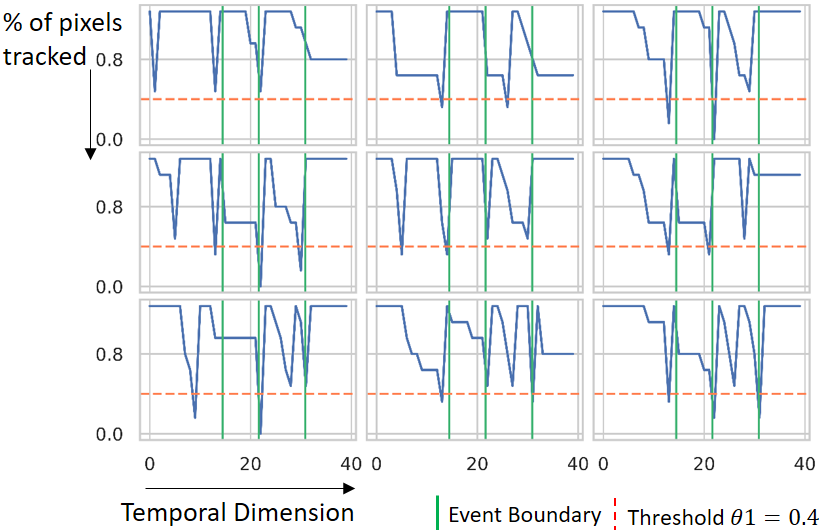}
	\caption{\textit{Pixel Tracking:} Visual representation of $3\times3$ patchwise pixel tracking along temporal dimension ($\theta_ 1=0.4$)}
	\label{fig:theta1}
\end{figure}
\setlength{\textfloatsep}{0.2cm}
\begin{figure}[tb]
	\centering
	\includegraphics[width=0.8\linewidth]{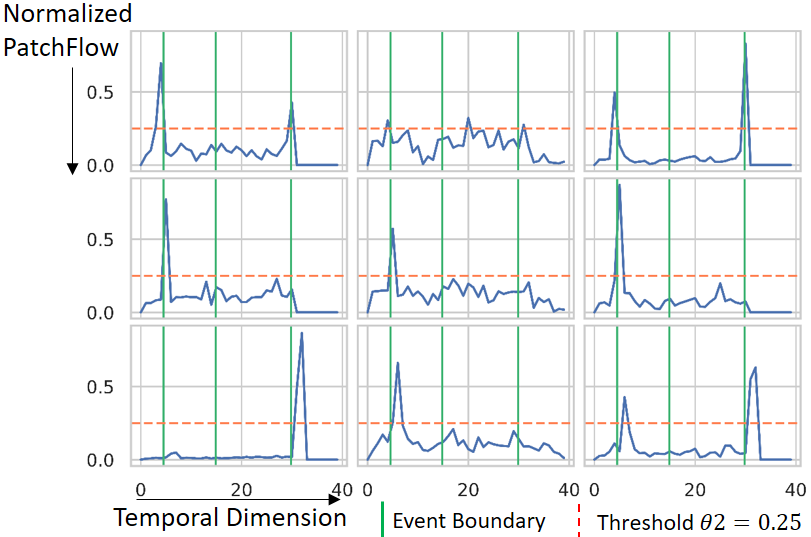}
	\caption{\textit{Flow Normalization:} Visual representation of normalized $3\times3$ patchwise max flow along temporal dimension ($\theta_2 = 0.25$)}
	\label{fig:theta2}
\end{figure}

\vspace{-1em}

\paragraph{Tuning Thresholds}\label{sec:tuningHyp}
$\theta_1$ and $\theta_2$ serve as thresholds that govern the behavior of the PT and FN algorithms, respectively. Additionally, $\theta_3$ is a threshold regulating neighboring boundaries for clustering in the refinement process.

As per the hypothesis, in PT, a notable decrease in pixel count during temporal tracking indicates pivotal event changes, such as changes in subjects or environment. Our empirical findings indicate that marking an event boundary with a drop exceeding 60\% ($\theta_1=0.4$) yields better performance. In Fig. \ref{fig:theta1}, it can be observed that the event boundaries are perfectly aligned with the trough. Likewise, in FN, our empirical observations indicate that marking a frame as an event boundary is effective when it contributes to over 25\% ($\theta_2=0.25$) of the overall normalized motion, signifying event changes in the video. Fig. \ref{fig:theta2} illustrates the alignment of peaks with event boundaries, visually validating our approach. The third threshold $\theta_3$ indicates the distance between two boundaries to consider them as belonging to the same cluster during refinement. Specifically, it is the Euclidean distance to split observations into clusters. We set $\theta_3 = 0.50$, i.e., twice the unit timestamp (time per frame) considered in our algorithm ($4$ FPS $\implies$ $1$ unit $= 1/4 = 0.25$).

All the experiments reported in Tables \ref{tab:KINETICS_RESULTS} and \ref{tab:TAPOS_RESULTS} are conducted in patchwise mode with $n_w = n_h = 5,  \theta_1 = 0.4, \theta_2 = 0.25,  \theta_3 = 0.5$. However, it is demonstrated in Section \ref{robust} that FlowGEBD is robust and insensitive to these thresholds. Qualitative results of FlowGEBD are presented in supplementary.

\subsection{Ablation Studies}

\paragraph{Effect of Sampling.} 
Table \ref{tab:AblationPT} illustrates the result of two sampling techniques. In random sampling, we uniformly sample the fixed fraction of pixels from each patch. The corner detection~\cite{ shi1994tomasi} looks for a significant change in pixel intensity in all directions. This sometimes results in the sampling of fewer corner pixels. Thus, we observe that random sampling of pixels gives better F1 scores for PT. It may be noted that the sampling method does not apply to the FN since it computes dense optical flow.

\vspace{-1em}
\paragraph{Effect of Spatial Granularity.}
As detailed in Section \ref{sec:PT_framewiseFlowGEBD} and \ref{sec:PT_patchwiseFlowGEBD}, we conclude from Table \ref{tab:AblationPT} that computing patchwise offers higher performance than processing the entire frame as a single unit. Further, by introducing Centroidal patches, we can capture event change at the intersection of the base patches, leading to a noticeable increase in F1.

\begin{table}[tbp]
	\begin{center}

		\resizebox{\columnwidth}{!}{%
			\begin{tabular}{c|cc|ccc|c}
				\hline
				   \multirow{2}{*}{\textbf{Method}}    & \multicolumn{2}{c|}{\textbf{Sampling} } & \multicolumn{3}{c|}{\textbf{Spatial Processing}} & \multirow{2}{*}{\textbf{F1@0.05}} \\ \cline{2-6}
				                              &   \textbf{Random}   &    \textbf{Corners}     &    \textbf{Framewise}    &   \textbf{Base Patch}    &  \textbf{Centroidal}   &                          \\ \hline
				                              &            &    \checkmark     & \checkmark &            &               &          0.492          \\
				                              & \checkmark &                   & \checkmark &            &               &          0.533          \\
				                              &            &    \checkmark     &            & \checkmark &               &          0.659          \\
				PT-\ref{algo:PixelTracking} & \checkmark &                   &            & \checkmark &               &          0.678          \\
				                              &            &    \checkmark     &            & \checkmark &  \checkmark   &          0.678          \\
				                              & \checkmark &                   &            & \checkmark &  \checkmark   &          0.702          \\ \hline
				     &            &                   & \checkmark &            &               &          0.486          \\
				            FN-\ref{algo:FlowBased}                  &     {\color{gray} NA}       &         {\color{gray} NA}          &            & \checkmark &               &          0.678          \\
				                              &            &                   &            & \checkmark &  \checkmark   &          0.691          \\ \hline
				          Ensembled           &     \checkmark       &         &            & \checkmark &  \checkmark   &     \textbf{0.713}      \\ \hline
			\end{tabular}%
		}
	\end{center}
	\caption{Effect of Sampling and Spatial Processing on Pixel Tracking (PT) and Flow-Normalized (FN) Algorithms}
	\label{tab:AblationPT}
\end{table}

\setlength{\textfloatsep}{0.1cm}
\setlength{\floatsep}{0.1cm}
\begin{table}[t]
	\begin{center}
		\resizebox{.7\columnwidth}{!}{%
			\begin{tabular}{c|c|c|c}
				\hline
				
				\multirow{2}{*}{$\mathcal{N}_g \impliedby (n_w = n_h)$}&      \multicolumn{3}{c}{\textbf{F1@0.05}}       \\ \cline{2-4}
				                                                                        & \textbf{PT-\ref{algo:PixelTracking}} & \textbf{FN-\ref{algo:FlowBased}} & \textbf{Ensembled}\\ \hline
				                                $n_w=3$                                 &     0.679     &   0.652   &  0.709   \\
				                                $n_w=4$                                 &     0.696     &   \textbf{0.694}   &  0.710   \\
				                                $n_w=5$                                 &     \textbf{0.702}     &   0.691   &  \textbf{0.713}   \\ \hline
			\end{tabular}%
		}
	\end{center}
	\caption{F1 score of the proposed method with respect to patch size on GEBD-Kinetics validation set}
	\label{tab:scalablePS}
\end{table}

\vspace{-1em}
\paragraph{Effect of Patch Size.}
Besides the spatial granularity, patch size is essential to predict the accurate event boundaries, as illustrated in Fig.  \ref{fig:flowchart}. We capture the effect of varying $n_w$ in Table \ref{tab:scalablePS}. A higher $\mathcal N_g$ (indicative of a smaller patch size) results in more candidate boundary sets, reducing the likelihood of missing uncommon boundaries. Moreover, it helps effectively trace events in tiny regions. We have determined the ideal value for our approach to be $n_w$ = $n_h$ = 5. Processing beyond a specific patch size can introduce noisy boundaries to the candidate pool, lowering the F1.

\subsection{Sensitivity Analysis of thresholds} \label{robust}

We conduct an extensive ablation study of thresholds and analyze the impact on the performance. We sample $\theta_1,\theta_2$ uniformly between $0.1$ to $0.9$, and for $\theta_3$, we vary it from $0.5$ to $3.0$ in steps of $0.5$.

\begin{figure}[t]

    \centering
    \includegraphics[width=0.8\linewidth]{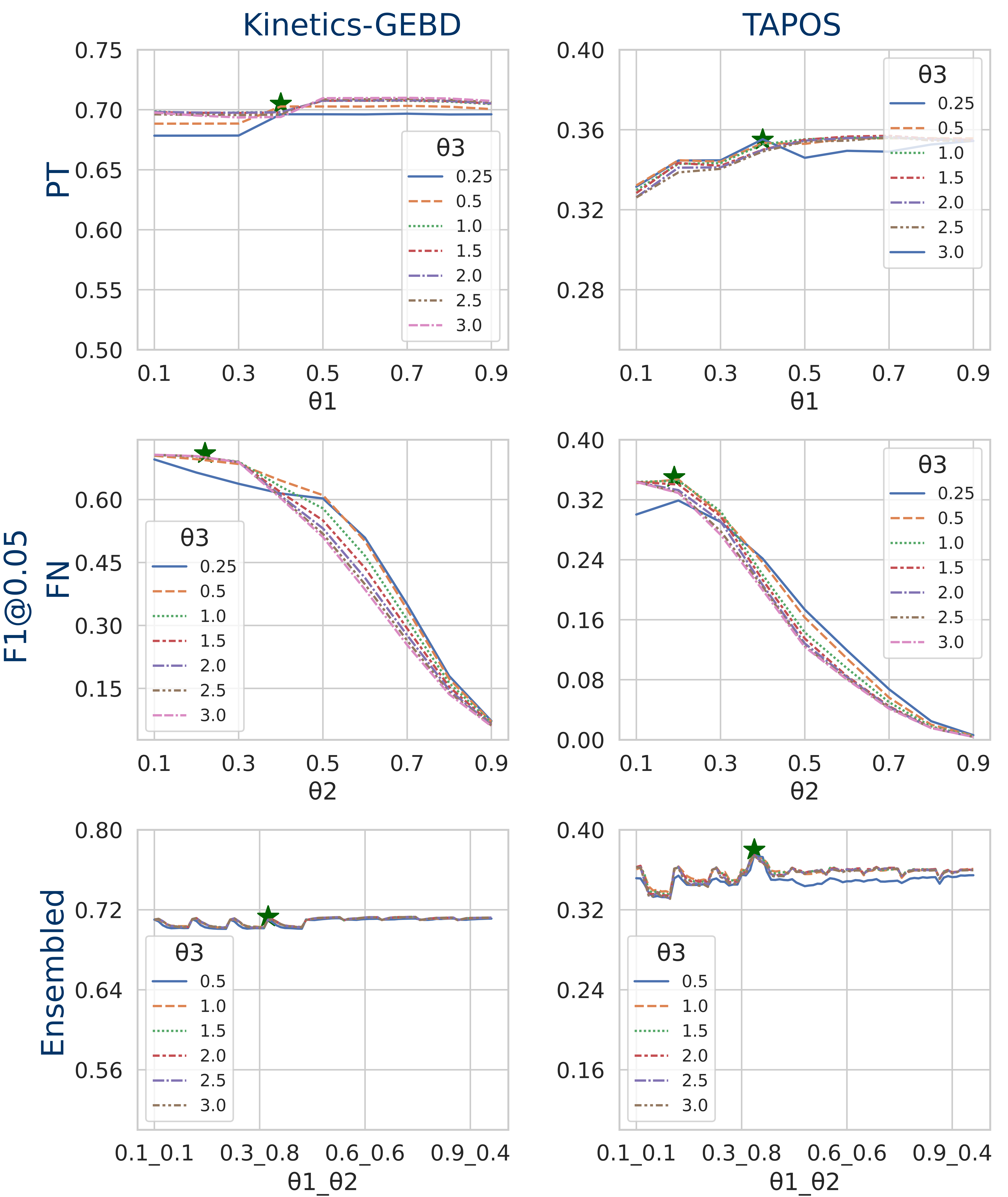}
    \caption{Sensitivity analysis of thresholds $\theta_1, \theta_2, \theta_3$. $\star$ marks the best F1@0.05 score.}
    \label{fig:AnalysisHP}

\end{figure}

Fig. \ref{fig:AnalysisHP} shows the analysis of sensitivity of the thresholds on Kinetics-GEBD and TAPOS datasets in patchwise ($n_w = 5$) mode. Our findings reveal that PT demonstrates robust performance across a wide range of threshold values ($\theta_1$ and $\theta_3$), consistently exhibiting the same trend for both the Kinetics and TAPOS datasets. In FN, as we increase the normalized flow threshold $\theta_2$, the number of detected boundaries will reduce gradually. The same effect is observed in Fig. \ref{fig:AnalysisHP}, where the gradual change in performance indicates relative stability and generalization on both datasets. 

The ensembled method shows the harmonious collaboration of PT and FN, attaining optimal F1 scores across all combinations ($9\times9\times6$). The mean standard deviations of F1@0.05 for PT, FN, and Ensembled on Kinetics-GEBD are $0.005$, $0.02$, and $0.0006$, respectively, while on TAPOS, they are $0.002$, $0.01$, and $0.01$. These findings highlight the robustness of FlowGEBD and its insensitivity to thresholds.

\setlength{\textfloatsep}{0.02cm}

\subsection{Time Complexity of FlowGEBD}
\vspace{-0.3em}
We theoretically assess the time complexity of FlowGEBD. From Algorithms \ref{algo:PixelTracking}, \ref{algo:FlowBased}, \ref{algo:filterBoundaries}, we observe that the time complexities (in patchwise mode) of pixel tracking, flow normalization, and temporal refinement are $O\left(L\mathcal{N}_g w_g h_g\left|\mathcal{P}_{\text{base}}\right|\right)$, $O(L\mathcal{N}_g w_g h_g)$, and $O(M \log M) \left(\equiv O(L \log L)\right)$, respectively. So, the overall time complexity is given by $O\left(L\mathcal{N}_g w_g h_g\left|\mathcal{P}_{\text{base}}\right| + L\mathcal{N}_g w_g h_g + L \log L\right)$. We can further simplify this to $ O\left(L\mathcal{N}_g w_g h_g\left|\mathcal{P}_{\text{base}}\right|\right)$. Since in pixel tracking, we use a sparse set of key pixels $\left(\text{i.e. }\left|\mathcal{P}_{\text{base}}\right| \ll w_g h_g\right)$, $\left|\mathcal{P}_{\text{base}}\right|$ is a fraction of $w$ and $h$. Additionally, $\mathcal{N}_g w_g h_g \propto wh$. So, we infer that the latency of FlowGEBD is directly proportional to $wh$ and $L$. Please consult the supplementary materials for the analysis of the inference time on sample videos.\looseness=-1

\setlength{\textfloatsep}{0.0cm}
\setlength{\floatsep}{0.0cm}
\begin{table}[t]
\centering
\resizebox{0.9\columnwidth}{!}{%
\begin{tabular}{c|c|c|c|c}
\hline
 & \textbf{Method} & \textbf{Params} & \textbf{Latency (ms)} & \textbf{F1@0.05} \\ \hline
 \multirow{9}{*}{ On GPU} & PC~\cite{shou2021generic}              & 23.5            & 46.4     &  0.625           \\
 & Gothe \etal \cite{gothe2023self}    & 6.79            & \textbf{1.2}       &      0.712      \\
 & Li \etal \cite{li2022end}        & $ \ge 34.6$          & 4.7      &     \textbf{0.743}        \\
 & PA~\cite{shou2021generic}               & 23.5            & 46.4          &   0.396     \\
 & UBoCo-Res50~\cite{kang2022uboco}     & $\ge 23.5 $          & $ \ge 46.4 $     &    0.703       \\
 & UBoCo-TSN~\cite{kang2022uboco}       & $\ge 90 $           & $\ge 90.2 $      &     0.702    \\ 
 \hline
 \multirow{5}{*}{ On CPU} 
 & Gothe \etal \cite{gothe2023self}    & 6.79            & 84.35        &    0.712       \\
 & SceneDetect~\cite{castellano2018pyscenedetect}      & {\color{gray} NA}               & 34.91       &   0.275       \\
 & Ours (PT \ref{algo:PixelTracking})           & {\color{gray} NA}               & \textbf{2.26}       &   0.702       \\
 & Ours (FN \ref{algo:FlowBased})         & {\color{gray} NA}               & \textbf{6.42}        &    0.691       \\
 & Ours (Ensembled)  & {\color{gray} NA}               &\textbf{ 6.5}    & \textbf{0.713} \\
 \hline
\end{tabular}%
}
\caption{Comparison of Latency with other methods and their F1@0.05 on Kinetics-GEBD validation set.}
\label{tab:latency}
\end{table}
\vspace{-1.0em}
\paragraph{Comparison of latency.} 
Table \ref{tab:latency} presents the latency per frame across different methods. Most of these methods employ ResNet-50 as their backbone, resulting in an average inference time of at least 46.4 ms per frame at a resolution of 160$\times$160 on a GPU \cite{gothe2023self}. In contrast, PT and FN exhibit considerably lower inference time, consuming 2.26 ms and 6.42 ms, respectively. The ensembled approach takes 6.5 ms on average without compromising the F1 score. The reported inference time is measured on a Samsung Galaxy S21 Ultra device with 12 GB RAM. Furthermore, the estimation of optical flow can be accelerated on GPU by utilizing NVIDIA Optical Flow SDK\cite{patait2019introduction}.\looseness=-1
\section{Conclusion and Discussion}

We introduce FlowGEBD, a non-parametric, unsupervised approach for generic event boundary detection. FlowGEBD comprises two independent algorithms, (i) Pixel Tracking and (ii) Flow Normalization, which can be deployed framewise or patchwise. FlowGEBD achieves the state-of-the-art results (Tables \ref{tab:KINETICS_RESULTS} and \ref{tab:TAPOS_RESULTS}) on the Kinetics-GEBD and TAPOS at a strict relative distance (F1@0.05). This demonstrates that the motion information acquired from an optical flow alone is sufficient and obviates the need for complex neural models to achieve high performance. We performed an extensive ablation study and threshold sensitivity analysis to demonstrate the robustness of the proposed method.\looseness=-1

However, since FlowGEBD does not incorporate spatial semantics (high-level DNN features), it is more suitable for GEBD rather than specific action/event localization. The same effect is observed in the evaluation of TAPOS. In future work, we will explore the bi-directional processing of each frame to improve the performance.

{\small
\bibliographystyle{ieee_fullname}
\bibliography{egbib}
}

\end{document}